\pdfoutput=1
\documentclass[11pt]{article}

\usepackage[preprint]{acl}

\usepackage{times}
\usepackage{latexsym}
\usepackage{tabularx}

\usepackage[T1]{fontenc}

\usepackage[utf8]{inputenc}

\usepackage{microtype}
\usepackage{booktabs}
\usepackage[table]{xcolor}
\usepackage{inconsolata}

\usepackage{graphicx}

\title{What Survives the Reset: Induction Circuits in Natural-Language Pre-Pretraining}

\author{
Elisabeth Fittschen\textsuperscript{1} \\
 \\\And
  Craig Messner\textsuperscript{1} \\
  \\\And
  Leshem Choshen\textsuperscript{2} \\
  \\\And  
  Tom Lippincott\textsuperscript{1} \\
\\
\hspace{-12cm}\textsuperscript{1}Johns Hopkins University, Department of Computer Science\\
\hspace{-12cm}\textsuperscript{2}Weizmann Institute of Science, Department of Mathematics and Computer Science\\
\hspace{-12cm}\small{
  \textbf{Correspondence:} \href{mailto:efittsc1@jh.edu}{efittsc1@jh.edu}
}
\\}

\begin{document}
\maketitle
\begin{abstract}
Pre-pretraining on synthetic data, such as formal languages, has been reported to improve the data efficiency of subsequent language model training, but what transfers, and for how long, is poorly understood. We address this by dissecting one clear transfer early in pretraining: even when embeddings, LM head and tokenizer are reset between stages, natural-language pre-pretraining, and to a lesser degree part-of-speech sequences, gives a much larger advantage over random initialization than the formal language Shuffle-Dyck. We show that this advantage is carried by a transferred induction circuit. After 30M words of pretraining, grafting only this circuit into a random model recovers $\sim$80\% of the full gain, re-drawing it in the pre-pretrained model removes almost all of it, and a hand-built circuit with no language-specific information recovers all of it. The advantage fades as the baseline forms its own circuit, and by 300M words of pretraining the best Shuffle-Dyck donor is the better prior. Our results provide a clear case of transferred computation, show that pre-pretraining benefits can vanish with longer pretraining, and suggest that understanding the mechanisms pre-pretraining instills may be necessary to predict benefit type and duration.
 
\end{abstract}
\begin{figure}[h]
  \centering
  \includegraphics[width=\columnwidth]{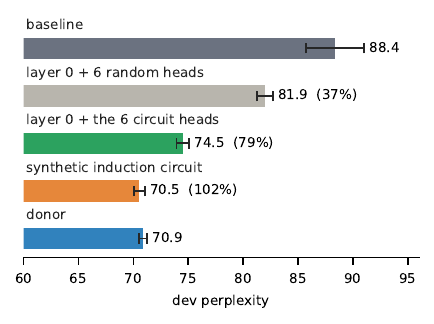}
    \caption{Identifying and transferring induction circuit weights after stage A is more than twice as effective compared to the same number of random heads.}
  \label{fig:abc}
\end{figure}
\section{Introduction}
\label{sec:intro}
 
Language models need large amounts of text: this creates a particular challenge for domains with data sparsity. \emph{Pre-pretraining} (\emph{stage-A}) is a short phase of training on a small, often synthetic corpus before standard pretraining (\emph{stage-B}).\footnote{We use "A" and "B" as a clearer distinction than multiple "pre"-prefixes, and acknowledge the irony in this footnote.} It can reduce how much text stage B needs: stage A on Shuffle-Dyck lets a 1B-parameter model reach its baseline loss with a third fewer tokens~\citep{hu2025circuits}.
 
Since synthetic corpora are cheap to generate and fully controllable, pre-pretraining is an attractive way to give a model useful inductive biases before it sees scarce data. Yet what transfers, and for how long, is poorly understood.
 
Why pre-pretraining helps is still debated. One view is that stage A teaches the model reusable computation: attention heads formed in stage A remain in use during stage B~\citep{hu2025circuits}, and transferring only the attention or only the MLP layers shows that the benefit sits in different components depending on the stage-A data~\citep{shinnick2025procedural,jiang2026procedural}. These results locate the benefit in whole layer types rather than in a specific mechanism, which leaves room for a simpler explanation: stage A may just leave the weights in a state that is easier to train from. Recent findings make this worth taking seriously: reported gains vary across model sizes, tokenizers and languages~\citep{riazhskykh2026instability}, and can vanish when the baseline starts from better-chosen random weights~\citep{anonymous2026rethinking}.
 
We address this by dissecting one clear transfer early in stage-B, where the benefit is large enough to take apart: after 10M words of stage B, natural-language pre-pretraining lowers perplexity by up to 45\%. Even when the embeddings, output head and tokenizer are reset between stages, natural-language pre-pretraining, and to a lesser degree part-of-speech (POS) sequences, gives a much larger head start than Shuffle-Dyck, a formal language chosen for its hierarchical structure. Since the reset removes everything tied to the stage-A vocabulary, the head start must come from the transformer body.
 
A natural candidate for a transferred computation is the induction circuit~\citep{elhage2021framework,olsson2022induction}. It completes repeated patterns by matching tokens on identity rather than meaning. The head start tracks it closely. Natural-language and POS donors form the circuit in stage A, pretraining reuses it at the donor’s own head positions, and their advantage closes once a randomly initialized model forms its own circuit. Shuffle-Dyck serves as a control: its donor forms no circuit, and its smaller advantage does not track induction.
 
Since tracking is only correlational, we also intervene (Figure \ref{fig:abc}). Grafting the circuit into an otherwise random model recovers most of the donor’s gain, about twice as much as random heads from the same layers, and re-initializing it in the donor removes almost all of it. Grafted weights could still carry latent language-specific information, so we also build the circuit by hand, from a few statistics measured on the donor but without copying any of its weights. Initializing a random model with it matches the full donor: what transfers is the circuit’s function, not knowledge of language, nor simply weights that are easier to train from.
 
The induction prior advantage is transient, suggesting that ranking of stage-A corpora at a single budget need not generalize. This also offers one explanation for why reported gains vary: a prior that the baseline soon learns on its own helps only at small budgets. And we argue that understanding the priors induced may be necessary to predict benefit consistency.
 
In summary, our contributions are:
\begin{itemize}
    \item Horizon dependence (\S\ref{sec:headstart}): the advantage of a stage-A corpus depends on the stage-B budget, so rankings obtained at a fixed budget may not generalize.
    \item A clear case of circuit transfer (\S\ref{sec:circuit}-\ref{sec:grafting}): the early advantage is carried by a specific induction circuit, not by generic properties of the pre-pretrained weights; grafting the donor’s induction circuit into an otherwise randomly initialized model recovers $\sim$80\% of the benefit of the full donor, while random heads from the same layers recover far less.
    \item Synthetic circuit initialization (\S\ref{sec:synthetic}): a circuit built by hand from statistics measured on random tokens yields comparable gains. The head start therefore does not require any knowledge leakage from the donor.
\end{itemize}

\section{Related Work}
\label{sec:related}

\paragraph{Pre-pretraining.}
Pre-pretraining trains a model on a small, often synthetic corpus before standard pretraining, so that it starts language learning with useful inductive biases. \citet{papadimitriou2020music} showed early on that LSTMs trained on music, code or parenthesis languages reach lower perplexity on natural language. Since then, transfer has been studied from artificial languages with controlled dependency structure~\citep{chiang2022transferability,ri2022artificial}, from priors distilled from Bayesian models of formal languages~\citep{mccoy2023priors}, from neural cellular automata~\citep{lee2026nca} and from procedural tasks~\citep{jiang2026procedural}.

\paragraph{Formal-language pre-pretraining.}
Our work builds most directly on \citet{hu2025circuits}. They hypothesize that a good pre-pretraining language is hierarchically structured and learnable by transformers in a length-generalizing way, which they operationalize as definability in C-RASP. Of four formal languages, $k$-Shuffle Dyck satisfies both criteria and transfers best to C4 for 160M-parameter Pythia models. \citet{mita2026languageacquisition} challenge the C-RASP criterion with MP-STRUCT, a formal language that is not C-RASP-definable yet transfers better than $k$-Shuffle Dyck.

\paragraph{Robustness of the gains.}
A replication of \citet{hu2025circuits} found that the gain did not consistently persist across model sizes, tokenizers and languages~\citep{riazhskykh2026instability}. When compute is matched, pre-pretraining does not beat spending the same compute on natural language. When it is added on top, its gains appear only at small natural-language budgets and depend on the baseline's initialization~\citep{anonymous2026rethinking}. 

\paragraph{Where transfer lives.}
\citeauthor{hu2025circuits}'s pruning shows that heads formed during pre-pretraining remain in use. Transferring only attention or only MLP layers shows that the transferable structure sits in different components depending on the data~\citep{shinnick2025procedural,jiang2026procedural}.
\paragraph{Induction heads.}
Induction heads complete repeated patterns. They form in a sharp transition early in training that coincides with gains in in-context learning~\citep{elhage2021framework,olsson2022induction}. Later work studies which subcircuits must form~\citep{singh2024induction}, which data statistics predict their formation~\citep{aoyama2025}, and their training dynamics~\citep{musat2025}. Architectural changes can make them easier to learn~\citep{xu2024}. We relate their formation in a randomly initialized model to the end of the pre-pretraining advantage, and install them through initialization.

\begin{figure*}[ht!]
  \centering
  \includegraphics[width=\textwidth]{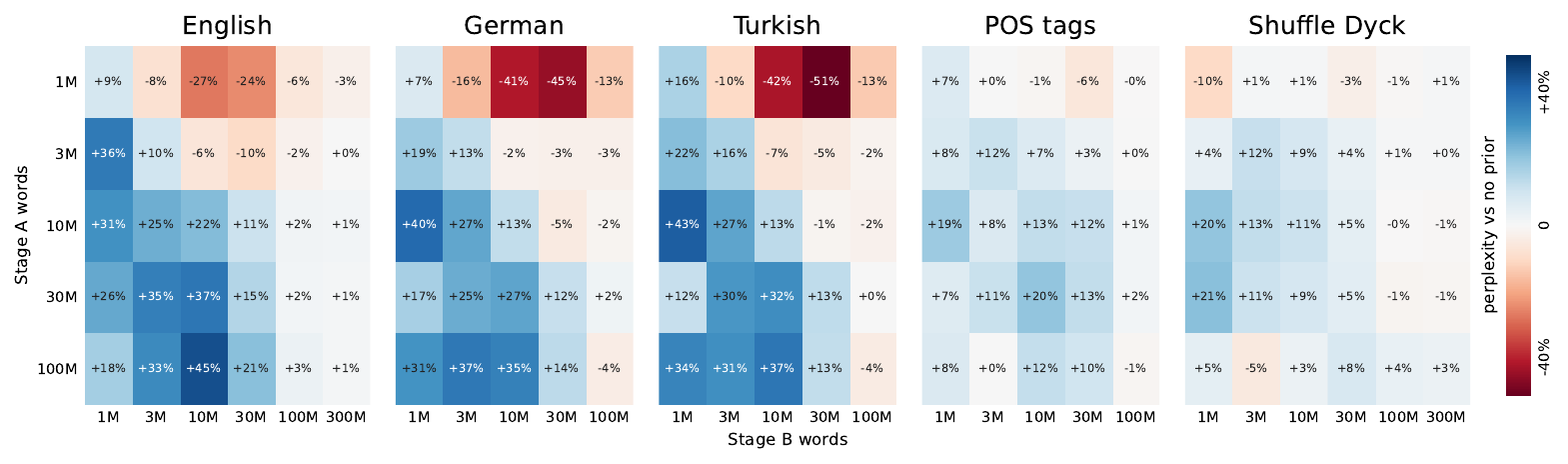}
  \caption{Perplexity reduction over the randomly initialized baseline, by stage-A budget (rows) and stage-B budget (columns), one panel per stage-A corpus. Each cell’s value is the average over three seeds.}
  \label{fig:heatmap}
\end{figure*}
\section{Experimental Framework}
\label{sec:framework}

\subsection{Model and training}
\label{sec:model}

All models share one 127M parameter 12-layer Llama architecture~\citep{touvron2023llama},  with 12 attention heads of dimension 64, hidden size 768, a SiLU-gated MLP of width 2688, RMSNorm and rotary position embeddings ($\theta = 10{,}000$). The Embeddings are untied, and attention is standard multi-head attention without grouped queries. We train with AdamW ($\beta_1 = 0.9$, $\beta_2 = 0.999$, $\epsilon = 10^{-6}$, weight decay 0.1), a cosine schedule with 10\% warmup that decays to 10\% of the peak learning rate, batches of $8 \times 2048$ tokens, bf16 and gradient clipping at 1.0. Each run trains a byte-level BPE tokenizer with 16k entries on its own data. Every configuration is run with three seeds, which set initialization, data order and tokenizer. A 100M-word run takes roughly 40 minutes on a single NVIDIA A100 or L40S GPU. This paper's runs, including learning-rate sweeps, total about 250 GPU hours. We report perplexity on held-out FineWeb text.

\paragraph{Corpora.}
We train donors on five corpora, three of natural language and two synthetic. 
\begin{itemize}
  \item \textbf{English, German and Turkish.} English text comes from FineWeb~\citep{penedo2024fineweb}, and German and Turkish text from its multilingual counterpart FineWeb-2~\citep{penedo2025fineweb2}. English stage-A and stage-B training data is disjoint.
  \item \textbf{Shuffle-Dyck.} We generate $k$-Shuffle Dyck as defined by \citet{hu2025circuits}: every opening bracket must be closed, but brackets of different types may cross. Following their settings, we use $k = 64$ bracket types (128 tokens), open a bracket with probability 0.5 and cap the nesting depth at 16.
  \item \textbf{POS.} We replace each word of the stage-A English FineWeb text by its Penn Treebank part-of-speech tag~\citep{marcus1993ptb}. This donor sees the word-class sequences of English without its words.
\end{itemize}
\subsection{Two-stage pipeline}
\label{sec:pipeline}

Every model passes through up to two stages. In stage~A, a donor is trained on a stage-A corpus. In stage~B, a model is trained on English text from the FineWeb sample-10BT shard~\citep{penedo2024fineweb}, split so that stage~B never overlaps with any stage-A text. Following \citet{hu2025circuits}, the optimizer state and learning-rate schedule are restarted between stages. The \emph{baseline} skips stage~A and starts stage~B from a random initialization.

Between the stages we reset the embeddings and the language-modeling head, that is, everything tied to the stage-A tokens. Each stage trains its own tokenizer on its own data, so an English donor's vocabulary never matches the stage-B tokenizer. Any benefit must result from some structure in the transformer body itself carrying over.
\begin{figure*}[t]
  \centering
  \includegraphics[width=\textwidth]{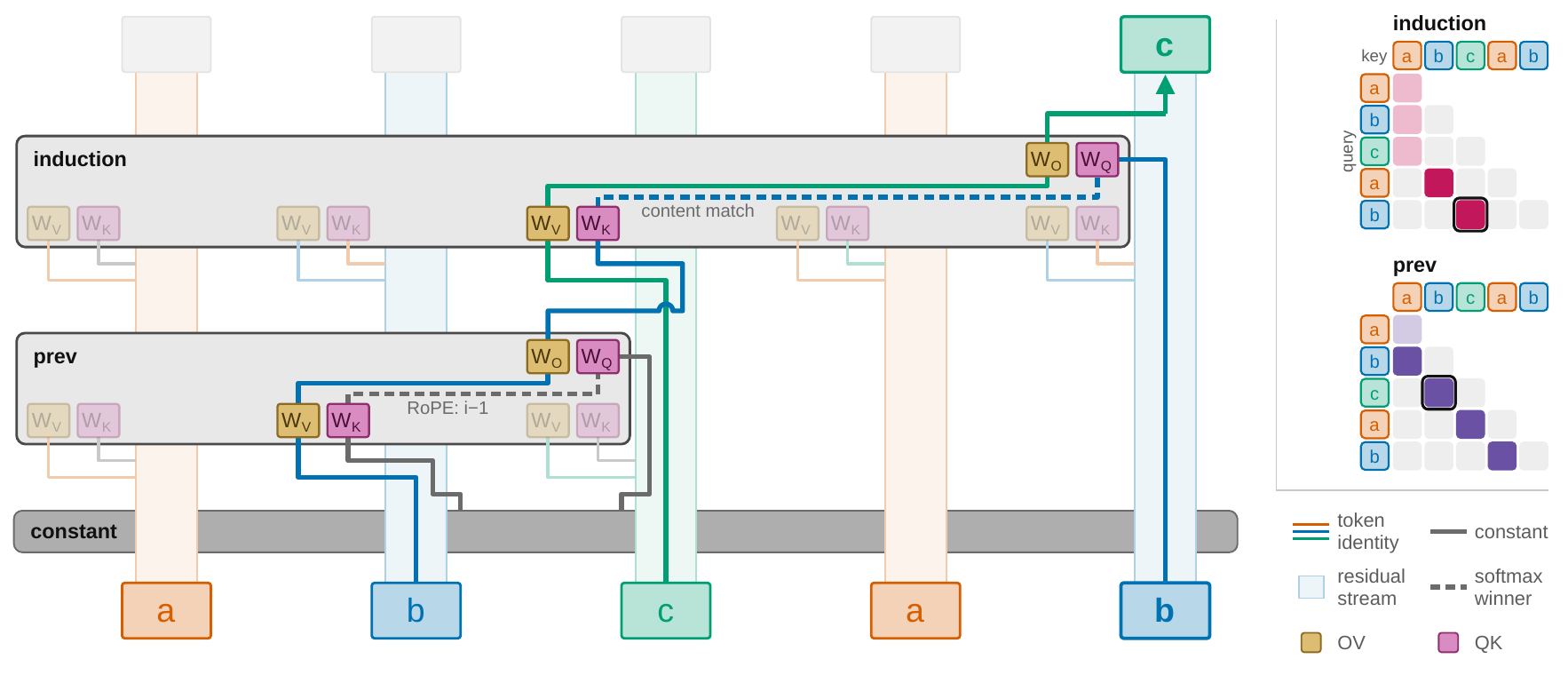}
  \caption{A visualization of the induction circuit. To show how information flows, the previous-token head is shown as it is at the query for token 2 (c) and the induction head state for query at position 4 (b). The right side displays the idealized attention patterns of the previous-token and induction head over the whole sequence.}
  \label{fig:induction}
\end{figure*}
\section{Stage A Gives a Head Start}
\label{sec:headstart}
 In stage A natural-language, and natural-language-like datasets such as POS bring the largest benefits early in stage B, and as its effects diminish the ranking of stage-A datasets changes. 

\subsection{Data Size Grid}
\label{sec:grid}

To explore the effect of stage-A and stage-B data size, we cross five stage-A  with five stage-B word budgets. Stage A is trained on English, German or Turkish text, on POS sequences, or on Shuffle-Dyck~\citep{hu2025circuits}, for 1M, 3M, 10M, 30M or 100M words. Each donor is then stage B trained on English for 1M, 3M, 10M, 30M or 100M words, about 1.4M to 141M tokens. Because stage B is always English, the German and Turkish donors test transfer across languages. We tune the learning rate of each stage for its own budget from a random initialization, so the stage-B settings are tuned for the baseline and not for each donor. Any bias therefore favors the baseline. To characterise longer stage B behavior we also run 300M stage B words for the English and Shuffle-Dyck donors. More details about hyperparameter tuning are in Appendix \ref{sec:appendix_hyperparameter_tuning}.

\subsection{Natural language leads early, and the lead fades}
\label{sec:lead}

Figure~\ref{fig:heatmap} shows the gain of every donor over the randomly initialized baseline. At small budgets, natural-language donors lead by a wide margin. After 10M words of stage B, the best English, German and Turkish donors lower perplexity by 45\%, 35\% and 37\%, against 20\% for POS and 11\% for Shuffle-Dyck. Most of the early benefit is therefore not English-specific knowledge.

The lead shrinks with the stage-B budget. After 100M words English donor advantage has sunken below Shuffle-Dyck and at 300M words Shuffle-Dyck advantage over English is above seed noise. The detailed perplexity and seed std results can be seen in Appendix \ref{sec:appendix_full_table}.

Small natural-language donors can hurt. A 1M-word donor helps after 1M words of stage B but worsens perplexity at every larger budget, in all three languages: by 24\% to 51\% after 10M and 30M words, and still by 6\% to 13\% after 100M. Why an under-trained donor is worse than none remains an open question.
\section{The Induction Circuit Carries the Head Start}
\label{sec:circuit}
\citet{hu2025circuits} show that heads formed during stage A remain in use. We examine attention heads and find that the natural-language head start tracks the formation of the induction circuit. And that donor-initialized models inherit the respective heads positions. 

\subsection{The Induction Circuit}
\label{sec:induction}
An induction circuit completes repeated patterns: having seen "a b c", it predicts "c" the next time "b" appears. Figure \ref{fig:induction} traces this on "a b c a b." First, a previous-token head attends from each position to the one before it, and writes that token’s identity into the current residual stream. To do so it needs to create a purely position-dependent attention pattern, it can do so by attending to a constant vector direction in the residual stream. Position 2 (c) thus also carries “preceded by b”. Second, an induction head in a later layer uses this information as its key. At the final b, its query matches the key at position 2, so it attends to  c and copies it to the output. The head’s query–key (QK) circuit decides where to look by content and its output–value (OV) circuit decides what to copy. No part of the circuit depends on token meaning.

\begin{figure*}[t]
  \centering
  \includegraphics[width=\textwidth]{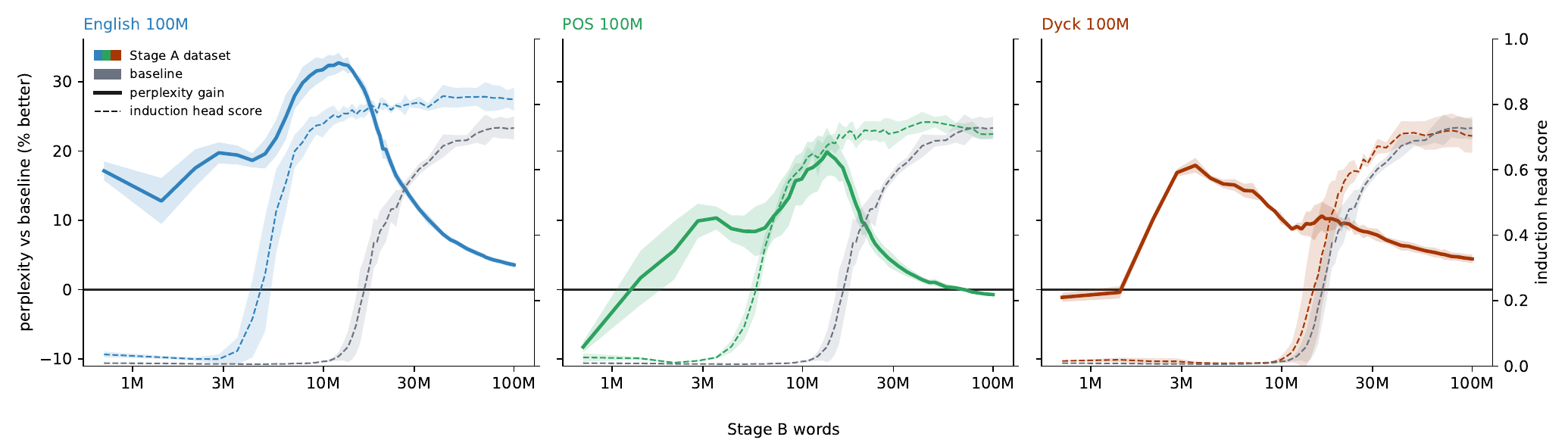}
  \caption{Perplexity difference between models initialized from 100M-word donors and the baseline during stage B. Dashed lines show the strongest prefix-matching score of the donor-initialized model (colored) and of the baseline (gray), on the right axis. Bands show the spread over three runs. %
  }
  \label{fig:tracking}
\end{figure*}

\subsection{Probing attention heads}
\label{sec:probe}
We use the standard random-token probe for induction heads~\citep{elhage2021framework,olsson2022induction}. We feed 16 sequences of 128 random tokens, each repeated once. A head's \emph{prefix-matching score} is its mean attention from a token in the repeat to the token that followed the same token's first occurrence. Its \emph{previous-token} and \emph{self} scores are its mean attention to the preceding and the current position. Chance is $1/128 \approx 0.008$, and a randomly initialized model scores about 0.006. Because the tokens are random, the probe reads attention geometry independently of what the embeddings encode, so it also applies to grafted models with fresh embeddings. We call a head an induction or previous-token head if its respective score exceeds 0.3 or 0.5. A visualization of these attention patterns can be seen in Figure \ref{fig:induction}.

\subsection{The early advantage tracks induction-circuit formation}
\label{sec:tracking}

Figure~\ref{fig:tracking} follows stage B from three 100M-word donors next to the baseline. It tracks the perplexity difference together with each model's strongest induction head. For the English and POS donors, the advantage moves in step with induction. The advantage rises as the donor-initialized model forms induction heads. It peaks at about 30\% (English) and 17\% (POS) while only that model has them, and closes as the baseline forms its own, between about 15M and 30M words. After 30M words the remaining gap shrinks slowly, to about 4\% for English and 0\% for POS at 100M words.

The Shuffle-Dyck donor behaves differently. Its model forms induction heads no earlier than the baseline, and its advantage peaks earlier, at about 15\% after 4M to 5M words, before any model has induction heads. A smaller but more constant advantage of about 5\% then persists to 100M words.

The advantages here do not perfectly follow advantages recorded in Figure \ref{fig:heatmap}, as these results represent advantages within a single 100M-word run.

\subsection{The donor's circuit is inherited}
\label{sec:inheritance}
\begin{table}[t]
\centering
\small
\setlength{\tabcolsep}{4pt}
\begin{tabular}{@{}ll rrrr@{}}
\toprule
donor & heads & shared & sites & Jaccard & $r$ \\
\midrule
baseline     & prev      & 0 & 10 & 0.14 & 0.21 \\
             & induction & 0 & 12 & 0.00 & 0.01 \\
\addlinespace[2pt]
Shuffle-Dyck & prev      & 1 &  8 & 0.34 & 0.51 \\
             & induction & 0 & 19 & 0.03 & 0.10 \\
\addlinespace[2pt]
POS     & prev      & 3 &  6 & 0.63 & 0.82 \\
             & induction & 4 &  5 & 0.87 & 0.97 \\
\addlinespace[2pt]
English      & prev      & 2 &  6 & 0.44 & 0.92 \\
             & induction & 3 &  4 & 0.83 & 0.97 \\
\bottomrule
\end{tabular}
\caption{Where the heads end up when the same donor body is trained with three stage-B seeds. \emph{Shared}: heads above threshold in all three seeds; \emph{Sites}: distinct heads above threshold in any seed; \emph{Jaccard}: mean overlap of the head sets between seed pairs; $r$: mean correlation of the 144 head scores between seed pairs.}
\label{tab:site_consistency}
\end{table}
Donors with an induction circuit pass it on at fixed positions. To test this, we train the same donor body with three stage-B seeds, which differ in embedding initialization, data order and tokenizer, and score all 144 heads on the random-token probe at the end of stage~B (Table~\ref{tab:site_consistency}).

Without a donor, head positions are not reproducible. Starting from the same randomly initialized body, the three seeds share no previous-token or induction head, and their induction-head sets do not overlap at all (Jaccard 0.00, $r = 0.01$). Neither the architecture nor the initialization determines where these heads form.

With an English or POS donor, induction heads land in nearly the same place in every seed (Jaccard 0.83--0.87, $r = 0.97$), and the heads shared across seeds are exactly the donor's stage-A heads. Previous-token sets overlap less (Jaccard 0.44--0.63) because individual seeds add further previous-token heads, but their score patterns stay highly correlated ($r \geq 0.82$). Stage B thus reuses the donor's circuit rather than building a new one.

Shuffle-Dyck behaves as expected for a donor without an induction circuit. It forms few previous-token heads and no induction heads during stage A, and passes on a single previous-token head (L1H11). Its induction heads form anew in each seed, across 19 different heads, consistent with its forming induction heads no earlier than the baseline (Figure \ref{fig:tracking}).

\section{The Circuit Carries the Head Start}
\label{sec:grafting}
The evidence so far is correlational. We now intervene directly, in two complementary ways. Grafting copies a group of components from the English 100M-word donor into an otherwise randomly initialized model. Re-drawing does the reverse: it re-initializes the group in the full donor and keeps everything else. After 30M words of stage B, we report the share of the donor's gain over baseline each model attains. 0\% is the baseline and 100\% the full donor.
\begin{table}[t]
\centering
\small
\setlength{\tabcolsep}{6pt}
\begin{tabular}{@{}l rr@{}}
\toprule
component group & grafted & re-drawn \\
\midrule
layer 0                                   & 14  & 30 \\
layer 0 + 2 previous-token heads              & 64  & 28 \\
layer 0 + 4 induction heads               & 39  & 14 \\
layer 0 + 2 prev + 4 induction heads      & 79  & 6 \\
\textit{layer 0 + 6 random heads}         & 37  & 28 \\
\midrule
2 previous-token heads                        & 31  & 87 \\
4 induction heads                         & 19  & 71 \\
2 prev + 4 induction heads                & 15  & 59 \\
\textit{6 random heads}                   & $-12$ & 97 \\
\midrule
full donor                                & 100 & 0 \\
\bottomrule
\end{tabular}
\caption{Sufficiency and necessity of the induction-circuit components of the English donor, as shares of the perplexity gain over the baseline (\%).
Negative values mean the model is worse than the baseline.}
\label{tab:graft_reverse}
\end{table}
\paragraph{The circuit is sufficient and necessary.}
Grafting layer~0 with the two previous-token and four induction heads recovers 79\% of the gain. Re-drawing the same components in the donor leaves 6\%, so the rest of the body carries almost none of the head start on its own. Six random heads from the same layers, grafted with layer~0, recover 37\%: the circuit heads add almost three times as much to layer~0 as random ones (65 vs.\ 23 percentage points).

\paragraph{Layer 0 is necessary but not sufficient.}
Grafted alone, layer 0 recovers only 14\% of the donor's advantage, yet re-drawing it in the donor removes 70\% (Table~\ref{tab:graft_reverse}). The rest of the circuit is built on it: without layer 0, the identified heads transfer poorly, at 31\% for the previous-token heads, 19\% for the induction heads, and only 15\% for both together, which is less than the previous-token heads alone. With layer 0, the same heads become effective. The two previous-token heads alone reach 64\%, and all six heads reach 79\%, far more than the sum of their separate contributions ($14 + 15 = 29\%$).

\begin{table}[t]
\centering
\small
\setlength{\tabcolsep}{5pt}
\begin{tabular}{@{}l rrr@{}}
\toprule
MLP0 in the donor & prev L1H8 & ind.\ L5H8 & copy loss \\
\midrule
intact                                   & 0.99 & 0.70 & 6.2 \\
MLP0 zeroed                              & 0.03 & 0.03 & 11.5 \\
MLP0 $\to$ its constant & 1.00 & 0.80 & 7.8 \\
layer-0 attn. zeroed   & 1.00 & 0.22 & 11.6 \\
\bottomrule
\end{tabular}
\caption{MLP0 can be replaced by the constant it writes(MLP0's mean output over text) and keep both heads and most of the copying.}

\label{tab:mlp0}
\end{table}

\paragraph{Layer 0 supplies the constant.}
Table~\ref{tab:mlp0} locates the constant in layer~0's MLP. Zeroing its output silences both the previous-token head L1H8 and the induction head L5H8, whereas zeroing layer-0 attention leaves the previous-token head intact. Replacing the output by its mean over positions, which keeps only the constant, preserves both the previous-token head fully and the induction head, removing the mean disables both.

\section{Building the Synthetic Circuit}
\label{sec:synthetic}

If the circuit's function is what transfers, it should be possible to install
it without any stage A. We write the circuit by hand into a randomly
initialized model, using only statistics measured on the English 100M-word
donor with the random-token probe. After 30M words of stage B, the synthetic
model reaches a dev perplexity of 70.5, against 70.9 for the full donor and
88.4 without a donor (medians over three seeds): 102\% of the donor's gain,
with every seed within the donor's seed spread. More details are documented in Appendix \ref{sec:appendix_synthetic}.

\begin{table}[t]
\centering
\small
\begin{tabular}{@{}lcccc@{}}
\toprule
 & \multicolumn{3}{c}{dev perplexity} & share \\
\cmidrule(lr){2-4}
initialization & s42 & s43 & s44 & (\%) \\
\midrule
random (baseline) & 92.4 & 86.0 & 88.4 & 0 \\
synthetic circuit & 70.5 & 70.1 & 71.3 & 102 \\
full donor        & 71.6 & 70.7 & 70.9 & 100 \\
\bottomrule
\end{tabular}
\caption{Dev perplexity after 30M words of stage B from each
initialization, and the share of the full donor's gain over random
initialization it attains.}
\label{tab:synthetic-results}
\end{table}

\paragraph{What the circuit needs.}
Four components, each placed in the donor's own heads.
\begin{enumerate}
  \setlength{\itemsep}{0pt}
  \item \textbf{Self heads} (L0H4, L0H7, L0H8) copy the current token into the
        residual stream.
  \item \textbf{MLP0} writes the same vector at every position, a constant
        that lets later heads attend by position alone.
  \item \textbf{Previous-token heads} (L1H8, L4H9) use it to attend one
        position back and pass that token forward.
  \item \textbf{Induction heads} (L4H2, L5H8, L7H1, L8H6) match the current
        token against the passed-forward tokens and copy the token that
        followed.
\end{enumerate}
Token identity travels through fixed 64-dimensional subspaces of the residual
stream, which we call \emph{codes}: the self heads write a
\emph{current-token code}, the previous-token heads a
\emph{previous-token code}. Every weight we write is added onto the random
initialization and keeps initialization-scale statistics, so the synthetic
model differs from a random one in structure, not in scale.

\paragraph{Background: rotary embeddings.}
RoPE splits each 64-dimensional query and key into 32 two-dimensional pairs,
each rotated by an angle proportional to the token's position, at its own
speed. \emph{Fast} pairs rotate a lot per position and tell nearby offsets
apart; \emph{slow} pairs barely rotate over short distances and act as a
position-free dot product. We call the 8 fastest pairs fast, the 12 slowest
slow and the rest \emph{middle}.

\paragraph{Background: the gated MLP.}
In the gated MLP, neuron $n$ reads the residual stream $x$ through a gate row
$g_n$ and an up row $u_n$ and writes through a down column $d_n$:
\begin{equation}
  \mathrm{MLP}(x) = \sum_n \mathrm{SiLU}(g_n \cdot x)\,(u_n \cdot x)\, d_n .
\end{equation}

\subsection{Layer 0: a token copy and a constant}
\label{sec:synthetic-layer0}

\paragraph{Self heads.}
\emph{Donor.} Three layer-0 heads attend almost only to the current token.
Their summed output, a copy of its embedding at about 32 times its norm, makes up 97\% of what MLP0 reads.

\emph{Synthetic.} Three heads at the same positions match each token to itself by content, not by position. Each value reads the token from one fixed code, and the output writes it into the current-token code. They reach self scores of 0.81, 0.45 and 0.46 (donor: 1.00, 0.66, 0.59), with output norms close to the donor's.

\paragraph{The MLP0 constant.}
Grafting showed that the previous-token heads read a constant written by
MLP0. No single neuron produces it.

\emph{Donor.}
\begin{enumerate}
  \setlength{\itemsep}{0pt}
  \item In some neurons the gate and up rows point in similar directions and
        the neuron fires with a positive mean; in equally many they point in
        opposite directions and it fires with a negative mean.
  \item Each down column leans along one shared direction, the
        \emph{carrier}, in proportion to that agreement.
  \item A negative mean times a lean away is a push toward, so 2{,}080 of
        2{,}688 neurons push the same way and the rest offset only 2\%.
\end{enumerate}
The sum is a constant of 0.38 of MLP0's output norm, spread over many
neurons; other seeds and donors follow the same law
(Table~\ref{tab:mlp0-law}).

\emph{Synthetic.} We impose this pattern on each neuron's random weights,
keeping every row's norm: part of each gate and up row is steered into the
current-token code, each neuron gets a random gate--up agreement with the
donor's spread, and its down column leans along a fixed carrier in
proportion. The constant reaches 0.50 of MLP0's output norm (donor: 0.38).

\subsection{Previous-token heads: fast pairs locate, slow pairs cancel}
\label{sec:synthetic-prev}

A head whose query and key are constant across positions can attend only by
offset, through RoPE: exactly what a previous-token head needs.

\emph{Donor.} L1H8 (score 0.99) and L4H9 (0.96) have queries and keys that
are almost constant and come mostly from the MLP0 carrier. Their keys use
the RoPE bands for two jobs:
\begin{itemize}
  \setlength{\itemsep}{0pt}
  \item \textbf{Fast pairs locate}: they make the logit peak at offset~1;
        without them the score drops to 0.08.
  \item \textbf{Slow pairs cancel}: barely rotating, they add the same
        negative logit to every key, pushing every offset but the first below
        zero even when the carrier's strength varies; without them the score
        drops to 0.75.
\end{itemize}

\emph{Synthetic.} Each head reads the carrier at its own layer. On fast
pairs, the key's phase leads the query's by one position's rotation, so the
logit peaks at offset~1. On slow pairs, query and key read the carrier with
opposite signs, adding a negative logit equal to minus the fast pairs' logit
at offset~2. The total is positive at offset~1, zero at offset~2 and negative
beyond, whatever the carrier's strength, since both terms scale with it.
Middle pairs are random.

\subsection{Induction heads: a content match on the slow pairs}
\label{sec:synthetic-induction}

An induction head looks for the earlier position whose previous token equals
the current one: a match by content, suited to the slow pairs.

\emph{Donor.} The four heads put 84--88\% of their query and key on the slow
pairs and 2--3\% on the fast ones. Their key arrives through the
previous-token heads, their query through MLP0 and the self head L0H4;
removing both previous-token heads drops L5H8 from 0.70 to 0.01.

\emph{Synthetic.} Each head's query reads the current-token code and its key
the previous-token code, on the slow pairs only and at a fixed strength,
with the component all tokens share removed so that only token identity is
matched. The value reads the current-token code and the output writes it
into a third code; other pairs are random. Before any training, the heads
score 0.74--0.81.

As in the non-synthetic donor, the synthetic circuit fixes where stage B
builds its induction heads, and they are recovered during stage B.
 
\section{Conclusion}
\label{sec:conclusion}
 
We asked what a pre-pretrained language model carries into pretraining, and for how long it matters. We studied 127M-parameter models pretrained on 1M to 300M words of English. Pre-pretraining on natural language gives them a head start: perplexity is 45\% lower than a random initialization after 10M words of pretraining, but only 1\% lower after 300M words.
 
Most of this head start is carried by one induction circuit, made of induction heads, the previous-token heads that supply their keys, and the layer-0 components that feed both. Grafted into a random model, this circuit recovers $\sim$80\% of the benefit of the full donor at 30M words pretraining. Built by hand from the donor's measured statistics, the circuit matches the full donor. 
 
Pre-pretraining datasets should therefore be compared across training horizons and judged by the inductive biases they install, since a bias helps only until a model would learn it from data. When the bias is known, it can be installed directly.

\section{Limitations}
\label{sec:limitation}

\paragraph{Scope of the decomposition.} Our decomposition, and the synthetic
circuit built from it, describes the English donor after 100M words of stage A
in a single 127M-parameter RoPE architecture. Longer-trained or larger models
may implement the same function differently. The previous-token mechanism
relies on RoPE and may not carry over to other position encodings.

\paragraph{One mechanism, short span.} We dissect a single mechanism. The
circuit matches on one preceding token and copies one token, so it explains
the completion of exact short repeats, not longer-range or fuzzy pattern
completion. The mechanisms behind other stage-A corpora, including
Shuffle-Dyck's early gain and its advantage at 300M words, are out of scope.

\paragraph{Size of the graft.} The grafted circuit includes all of layer 0
(all 12 heads and MLP0), so the amount of parameters transferred is about 9.7M parameters: 7.7\% of the model and 9.5\% of the transformer body.

\paragraph{Tokenization.} Each stage trains its own tokenizer on its own data,
so perplexity is per token of a budget-specific tokenizer. Comparisons within a
column of Table~\ref{tab:ab_cross_seeds_ppl} are valid, absolute perplexities across
stage-B budgets are not directly comparable.

\paragraph{Experimental scale.} All results use three seeds, English stage-B
data and perplexity as the only metric, we do not evaluate in-context learning
or downstream tasks. Stage-B learning rates are tuned for the random baseline,
which may disadvantage donors. Grafting and the synthetic circuit are tested
at one stage-B budget (30M words). The ranking change
at 300M words rests on a single shuffle-dyck donor (100M words). However this donors out-performance of other shuffle-dyck donors occurs consistently in the stage-B 30M 100M and 300M runs, and it is not an outlier. 
\bibliography{custom}

\appendix

\section{Hyperparameter Tuning}
\label{sec:appendix_hyperparameter_tuning}
Table \ref{tab:stageA_lr} shows the hyperparameter tuning results used in the stage-A x stage-B table. Each setting was hyperparameter tunes on random initialization and optimized for performance on its own task. Hyperparameter tuning started with a lr grid of 3, 10, 30 $\times 10^{-4}$, expanded the grid whenever the optimum lies on an edge, and tests intermediate learning rates ($2, 5, 20 \times 10^{-4}$) around the current optimum.
\begin{table*}[t]
\centering
\small
\begin{tabular}{@{}llrrrr>{\columncolor{black!8}}rr>{\columncolor{black!8}}rr>{\columncolor{black!8}}rrr@{}}
\toprule
& & \multicolumn{11}{c}{peak learning rate ($\times 10^{-4}$)} \\
\cmidrule(lr){3-13}
corpus & words & 0.3 & 0.5 & 1 & 2 & 3 & 5 & 10 & 20 & 30 & 50 & 100 \\
\midrule
English & 1M   &     &     &       &       & 1835  &       & 1699  & \textbf{1537} & 1577  & 1550 & 1675 \\
        & 3M   &     &     &       &       & 741   & 683   & \textbf{643} & 714 & 785 &  &  \\
        & 10M  &     &     &       &       & 271   & \textbf{245}  & 247  & 326 & 440 &  &  \\
        & 30M  &     &     & 173   & 121   & 106   & \textbf{94.2} & 124  &     & 330 &  &  \\
        & 100M &     &     &       &       & 50.4  & \textbf{46.6} & 48.1 & 168 & 261 &  &  \\
\addlinespace[2pt]
Shuffle-Dyck    & 1M   & 257 & 257 & \textbf{257} & 257 & 257 &      & 258  &     & 258 &  &  \\
        & 3M   &     &     & 234   & 230   & \textbf{229}  & 233  & 252  &     & 257 &  &  \\
        & 10M  &     &     & 111   & 105   & 84.2  & \textbf{69.4} & 102  &     & 133 &  &  \\
        & 30M  &     &     & 68.2  & 60.0  & 58.9  & \textbf{57.9} & 59.0 &     & 147 &  &  \\
        & 100M &     &     & 57.1  & 56.2  & \textbf{56.1} & 56.1 & 56.5 &     & 155 &  &  \\
\addlinespace[2pt]
POS     & 1M   &     &     & 11.94 & 11.73 & \textbf{11.69} & 12.00 & 12.43 &       & 17.34 &  &  \\
        & 3M   &     &     & 10.64 &       & 9.81  & \textbf{9.63} & 9.77  & 10.13 & 10.75 &  &  \\
        & 10M  &     &     & 8.31  & 8.13  & \textbf{8.08} & 8.10  & 8.32  &       & 9.15  &  &  \\
        & 30M  &     &     & 7.65  & 7.48  & 7.37  & \textbf{7.35} & 7.75  &       & 9.07  &  &  \\
        & 100M &     &     & 7.02  & 6.92  & 6.88  & \textbf{6.88} & 7.09  &       & 8.76  &  &  \\
\bottomrule
\end{tabular}
\caption{Stage-A learning-rate sweep: dev perplexity on each corpus's own held-out stream, mean of 3 seeds. Shaded: initial grid; further rates extended edge winners, then flanked each winner at log midpoints. Bold: selected, blind to stage B. Blank: not run.}
\label{tab:stageA_lr}
\end{table*}

\section{Full Table}
\label{sec:appendix_full_table}
Table \ref{tab:ab_cross_seeds_ppl} shows all stage-A x stage-B mean results over three seeds, and also includes the per setting standard deviation. The baseline and english 100M donor runs were rerun in the graft and synthetic circuit sections, and due to non determinism their medians/means differ slightly.

At 300M words of Stage B data the 100M-word Shuffle Dyck donor reaches a lower perplexity than the 100M-word English donor (32.2 vs 32.9, better in every seed, $p = 0.01$, paired over seeds), and it keeps a 3\% margin over the baseline
(33.2) where the English donor's margin has shrunk to within noise (0.8\%).

\begin{table*}[t]
\centering
\small
\setlength{\tabcolsep}{4pt}
\begin{tabular}{@{}lrrrrrr@{}}
\toprule
Stage A & \multicolumn{6}{c}{Stage B words} \\
\cmidrule(lr){2-7}
words & 1M & 3M & 10M & 30M & 100M & 300M \\
\midrule
\multicolumn{7}{@{}l}{\emph{English}} \\
1M & $1428.4 \pm 73.2$ & $707.2 \pm 34.2$ & $317.9 \pm 3.8$ & $112.1 \pm 2.8$ & $49.4 \pm 0.7$ & $34.1 \pm 0.8$ \\
3M & $1000.0 \pm 40.1$ & $587.8 \pm 30.6$ & $265.6 \pm 13.4$ & $99.1 \pm 5.6$ & $47.6 \pm 0.2$ & $33.2 \pm 0.3$ \\
10M & $1083.8 \pm 109.2$ & $490.9 \pm 29.9$ & $195.0 \pm 0.8$ & $80.4 \pm 0.6$ & $45.7 \pm 0.0$ & $32.8 \pm 0.2$ \\
30M & $1158.4 \pm 197.2$ & $425.9 \pm 8.9$ & $157.2 \pm 6.1$ & $76.5 \pm 0.4$ & $45.9 \pm 0.4$ & $32.9 \pm 0.5$ \\
100M & $1296.3 \pm 227.0$ & $441.9 \pm 14.9$ & $137.5 \pm 2.4$ & $70.9 \pm 0.9$ & $45.3 \pm 0.5$ & $32.9 \pm 0.4$ \\
\addlinespace[2pt]
\multicolumn{7}{@{}l}{\emph{German}} \\
1M & $1456.6 \pm 76.5$ & $763.0 \pm 110.5$ & $353.8 \pm 15.6$ & $130.7 \pm 10.7$ & $52.7 \pm 0.5$ & -- \\
3M & $1279.9 \pm 72.8$ & $573.7 \pm 49.1$ & $255.6 \pm 6.2$ & $92.6 \pm 3.6$ & $47.9 \pm 0.4$ & -- \\
10M & $944.9 \pm 36.8$ & $481.4 \pm 46.9$ & $218.5 \pm 6.3$ & $94.4 \pm 0.6$ & $47.4 \pm 0.1$ & -- \\
30M & $1306.5 \pm 91.2$ & $489.2 \pm 26.9$ & $182.4 \pm 3.9$ & $79.6 \pm 1.9$ & $45.7 \pm 1.0$ & -- \\
100M & $1091.2 \pm 89.2$ & $416.4 \pm 25.6$ & $163.4 \pm 22.9$ & $77.9 \pm 0.6$ & $48.5 \pm 0.2$ & -- \\
\addlinespace[2pt]
\multicolumn{7}{@{}l}{\emph{Turkish}} \\
1M & $1318.8 \pm 22.1$ & $722.3 \pm 23.1$ & $355.7 \pm 32.0$ & $136.5 \pm 7.3$ & $52.8 \pm 2.3$ & -- \\
3M & $1230.8 \pm 124.3$ & $550.3 \pm 24.7$ & $267.1 \pm 10.5$ & $94.5 \pm 2.2$ & $47.4 \pm 0.7$ & -- \\
10M & $903.9 \pm 81.3$ & $478.6 \pm 17.1$ & $217.0 \pm 3.8$ & $91.5 \pm 2.0$ & $47.7 \pm 0.2$ & -- \\
30M & $1389.6 \pm 11.1$ & $459.5 \pm 11.2$ & $170.0 \pm 12.4$ & $78.3 \pm 1.7$ & $46.5 \pm 0.4$ & -- \\
100M & $1041.0 \pm 113.5$ & $450.5 \pm 9.5$ & $157.1 \pm 4.5$ & $78.9 \pm 2.7$ & $48.7 \pm 0.5$ & -- \\
\addlinespace[2pt]
\multicolumn{7}{@{}l}{\emph{POS tags}} \\
1M & $1462.9 \pm 110.2$ & $655.5 \pm 28.9$ & $252.8 \pm 6.4$ & $95.5 \pm 3.4$ & $46.7 \pm 0.3$ & -- \\
3M & $1452.6 \pm 248.7$ & $580.6 \pm 45.1$ & $232.6 \pm 4.1$ & $87.6 \pm 3.6$ & $46.6 \pm 0.7$ & -- \\
10M & $1281.0 \pm 27.4$ & $606.5 \pm 17.5$ & $218.7 \pm 3.6$ & $79.6 \pm 1.1$ & $46.0 \pm 0.3$ & -- \\
30M & $1469.8 \pm 105.9$ & $582.4 \pm 24.1$ & $199.5 \pm 4.0$ & $78.6 \pm 1.0$ & $45.9 \pm 0.7$ & -- \\
100M & $1453.8 \pm 170.7$ & $655.8 \pm 46.0$ & $219.4 \pm 7.1$ & $81.1 \pm 1.2$ & $47.2 \pm 0.2$ & -- \\
\addlinespace[2pt]
\multicolumn{7}{@{}l}{\emph{Shuffle Dyck}} \\
1M & $1734.1 \pm 186.7$ & $650.4 \pm 18.1$ & $248.6 \pm 2.9$ & $92.8 \pm 2.2$ & $47.0 \pm 0.2$ & $33.0 \pm 0.5$ \\
3M & $1504.1 \pm 54.4$ & $578.7 \pm 18.9$ & $228.8 \pm 5.0$ & $86.9 \pm 1.0$ & $46.3 \pm 0.7$ & $33.1 \pm 0.4$ \\
10M & $1255.1 \pm 165.9$ & $572.8 \pm 30.6$ & $222.6 \pm 4.4$ & $85.8 \pm 1.1$ & $46.8 \pm 0.3$ & $33.4 \pm 0.3$ \\
30M & $1237.7 \pm 107.5$ & $584.4 \pm 14.1$ & $229.0 \pm 7.5$ & $85.3 \pm 1.2$ & $47.2 \pm 0.3$ & $33.5 \pm 0.4$ \\
100M & $1494.9 \pm 76.9$ & $688.1 \pm 46.3$ & $242.6 \pm 4.4$ & $82.9 \pm 1.0$ & $44.9 \pm 0.2$ & $32.2 \pm 0.3$ \\
\addlinespace[2pt]
\midrule
\multicolumn{7}{@{}l}{\emph{baseline dev perplexity}} \\
& $1573.5 \pm 149.5$ & $656.2 \pm 16.8$ & $250.4 \pm 4.3$ & $90.2 \pm 3.4$ & $46.6 \pm 0.6$ & $33.2 \pm 0.7$ \\
\bottomrule
\end{tabular}
\caption{Dev perplexity of every cell of the Stage A $\times$ Stage B cross: mean $\pm$ standard deviation over the three seeds. The last row is the baseline (no Stage A). -- marks a cell that was not run. The heatmap shows $100\,(1 - \mathrm{cell}/\mathrm{baseline})$ of the seed means.}
\label{tab:ab_cross_seeds_ppl}
\end{table*}

\section{Constructing the Synthetic Circuit}
\label{sec:appendix_synthetic}

This appendix gives the measurements and construction behind
the synthetic circuit.

\paragraph{Terms.}
The \emph{energy} of a vector in a subspace is the share of its squared norm
lying in that subspace. The \emph{lean} of a down column is its component
along the carrier, as a fraction of the column's norm.
$\rho_n$ is the cosine between neuron $n$'s gate and up rows, measured inside
the 64 directions that carry MLP0's input.

\subsection{Self heads}
\label{app:self}

\emph{Donor.} Self scores are 1.00, 0.66 and 0.59, with output norms 17.7,
4.2 and 3.9 against an embedding norm of 0.75. The summed attention output
(norm 24.0) makes up 97\% of MLP0's input. It compresses the embedding's 63
effective dimensions to 25.

\emph{Synthetic.} Each head's query and key projections are two random
64-row matrices (frames) whose corresponding rows have cosine 0.71. The query
and key of the same token therefore align, so the head matches the current
token by content rather than position. The value reads one fixed
64-dimensional code subspace and the output writes a second.

\emph{Result.} Self scores 0.81, 0.45 and 0.46; output norms 18.8, 4.0 and
4.2.

\subsection{The MLP0 constant}
\label{app:mlp0}

\emph{Donor.}
\begin{itemize}
  \item The gate--up cosine $\rho_n$ has spread $\sigma_\rho = 0.24$, against
        0.12 at random initialization; both signs are equally common.
  \item A neuron's mean activation rises with $\rho_n$ (slope 0.68).
  \item Its down column's lean rises with $\rho_n$ (slope 0.25). Lean and
        mean activation correlate at $r = 0.80$.
  \item 2{,}080 of 2{,}688 neurons push the same way; the rest contribute 2\%.
  \item The constant is 0.38 of MLP0's output norm (65). 132 neurons give
        half of it, 649 give 90\%.
\end{itemize}

\emph{Synthetic.} Three operations per neuron, keeping every row and its
norm:
\begin{enumerate}
  \item Rotate 40\% of each gate and up row's energy into the self heads'
        code subspace (donor: 41\%).
  \item Set the up row's component in that subspace at cosine
        $\rho_n \sim \mathcal{N}(0, 0.24)$ to the gate row's.
  \item Tilt the down column along one fixed carrier direction by
        $0.24\,\rho_n$ of its norm.
\end{enumerate}

\emph{Result.} Cosine spread 0.24, lean slope 0.25, mean-activation slope
0.74. The constant is 0.50 of an output norm of 35 (donor: 0.38 of 65); 288
neurons give half of it, 994 give 90\%.

\begin{table}[t]
\centering
\small
\setlength{\tabcolsep}{4pt}
\begin{tabular}{@{}lccccc@{}}
\toprule
donor & $\sigma_\rho$ & lean & $r$ & const & against \\
\midrule
random init       & 0.12 & 0.05 & --   & 0.19 & --  \\
English, seed 42  & 0.24 & 0.24 & 0.80 & 0.38 & 2.3 \\
English, seed 43  & 0.27 & 0.28 & 0.79 & 0.47 & 1.7 \\
English, seed 44  & 0.26 & 0.23 & 0.82 & 0.41 & 2.4 \\
German            & 0.28 & 0.25 & 0.73 & 0.43 & 2.8 \\
Turkish           & 0.29 & 0.24 & 0.78 & 0.44 & 1.8 \\
POS               & 0.37 & 0.26 & 0.63 & 0.59 & 2.1 \\
synthetic circuit & 0.24 & 0.25 & --   & 0.50 & --  \\
\bottomrule
\end{tabular}
\caption{The MLP0 constant law across 100M-word donors. $\sigma_\rho$: spread
of the gate--up cosine inside MLP0's input subspace; lean: slope of the down
column's lean on that cosine; $r$: correlation between a column's lean and
its neuron's mean activation; const: constant as a fraction of MLP0's output
norm; against: share of the constant (\%) carried by neurons pushing the
other way.}
\label{tab:mlp0-law}
\end{table}

\subsection{Previous-token heads}
\label{app:prev}

\emph{Donor.} L1H8 scores 0.99 and L4H9 0.96. Their queries and keys are
97--98\% constant across tokens and come 93--94\% from the MLP0 carrier. For
a query and key that are constant across tokens, RoPE makes the logit a
function of the offset $\Delta$ alone:
\begin{equation}
  \ell(\Delta) = \sum_{p=1}^{32} A_p \cos(\theta_p \Delta - \varphi_p),
  \label{eq:rope-logit}
\end{equation}
with $\theta_p = 10000^{-2(p-1)/64}$. Here $A_p$ is the product of the query
and key amplitudes on pair $p$, and $\varphi_p$ the phase by which the key
leads the query. Pairs 1--8 are fast, 9--20 middle, 21--32 slow.

The donor's key puts 27\% of its energy on the fast pairs, 17\% on the middle
ones and 55\% on the slow ones, with its phase locked on the fast pairs.
Zeroing the fast pairs drops the head to 0.08; zeroing the slow pairs drops
it to 0.75.

\emph{Synthetic.} Each head reads the carrier at its own layer.
\begin{itemize}
  \item \textbf{Fast pairs:} $\varphi_p = \theta_p$, so each term is
        $A_p \cos(\theta_p(\Delta - 1))$, maximal at $\Delta = 1$.
  \item \textbf{Slow pairs:} query and key read the carrier with opposite
        signs. Since $\theta_p \Delta \approx 0$ over short offsets, each
        term is a near-constant negative; amplitudes are set so their sum
        equals $-\ell_{\mathrm{fast}}(2)$.
  \item \textbf{Middle pairs:} random amplitude and phase.
\end{itemize}
Ignoring the middle pairs,
$\ell(\Delta) \approx \ell_{\mathrm{fast}}(\Delta) - \ell_{\mathrm{fast}}(2)$:
positive at $\Delta = 1$, zero at $\Delta = 2$, negative beyond. Every $A_p$
scales with the square of the carrier's amplitude, so this sign pattern holds
for any amplitude.

\emph{Result.} Scores 0.97 and 0.99; 0.68 without the slow-pair term.

\subsection{Induction heads}
\label{app:induction}

\emph{Donor.} The four heads put 84--88\% of their query and key energy on
the 12 slow pairs and 2--3\% on the fast ones. For L5H8, the key arrives 53\%
via L1H8 and 35\% via L4H9; the query 51\% via MLP0 and 18\% via L0H4.
Zeroing both previous-token heads drops L5H8 from 0.70 to 0.01.

\emph{Synthetic.} The query reads the current-token code and the key the
previous-token code, each through the 24 directions of its code that carry
the least energy. This excludes the high-energy component that all tokens
share. The match uses the 12 slow pairs and is scaled to a mean logit of 12
for matching tokens, at each head's own layer. Other pairs carry random
amplitude and phase. The value reads the current-token code; the output
writes it along a third subspace.

\emph{Result.} Before any training the heads score 0.77, 0.81, 0.79 and 0.74.

\end{document}